\documentclass[10pt,twocolumn,letterpaper]{article}

\usepackage[pagenumbers]{cvpr} 

\usepackage{graphicx}
\usepackage{amsmath}
\usepackage{amssymb}
\usepackage{booktabs}
\usepackage{bm}

\usepackage[pagebackref,breaklinks,colorlinks]{hyperref}

\makeatletter
\def\blfootnote{\gdef\@thefnmark{}\@footnotetext}
\makeatother
\usepackage[capitalize]{cleveref}
\crefname{section}{Sec.}{Secs.}
\Crefname{section}{Section}{Sections}
\Crefname{table}{Table}{Tables}
\crefname{table}{Tab.}{Tabs.}

\def\confName{CVPR}
\def\confYear{2023}

\begin{document}

\title{Design Booster: A Text-Guided Diffusion Model for Image Translation with Spatial Layout Preservation}

\author{%
\hspace{4mm}Shiqi Sun*%
\hspace{4mm}Shancheng Fang*%
\hspace{4mm}Qian He%
\hspace{4mm}Wei Liu%
\vspace{1.7mm}\\ByteDance Ltd, Beijing, China\\%
\hspace{4mm}{\tt\small sunshiqi.2333@bytedance.com}
\hspace{4mm}{\tt\small fangshancheng.lh@bytedance.com}
\hspace{4mm}{\tt\small heqian@bytedance.com}\\
\hspace{4mm}{\tt\small liuwei.jikun@bytedance.com}
}

\twocolumn[{
\renewcommand\twocolumn[1][]{#1}
\maketitle
\vspace*{-10mm}
\begin{center}
    \label{mainmap}
    \centering
    \includegraphics[width=1\textwidth]{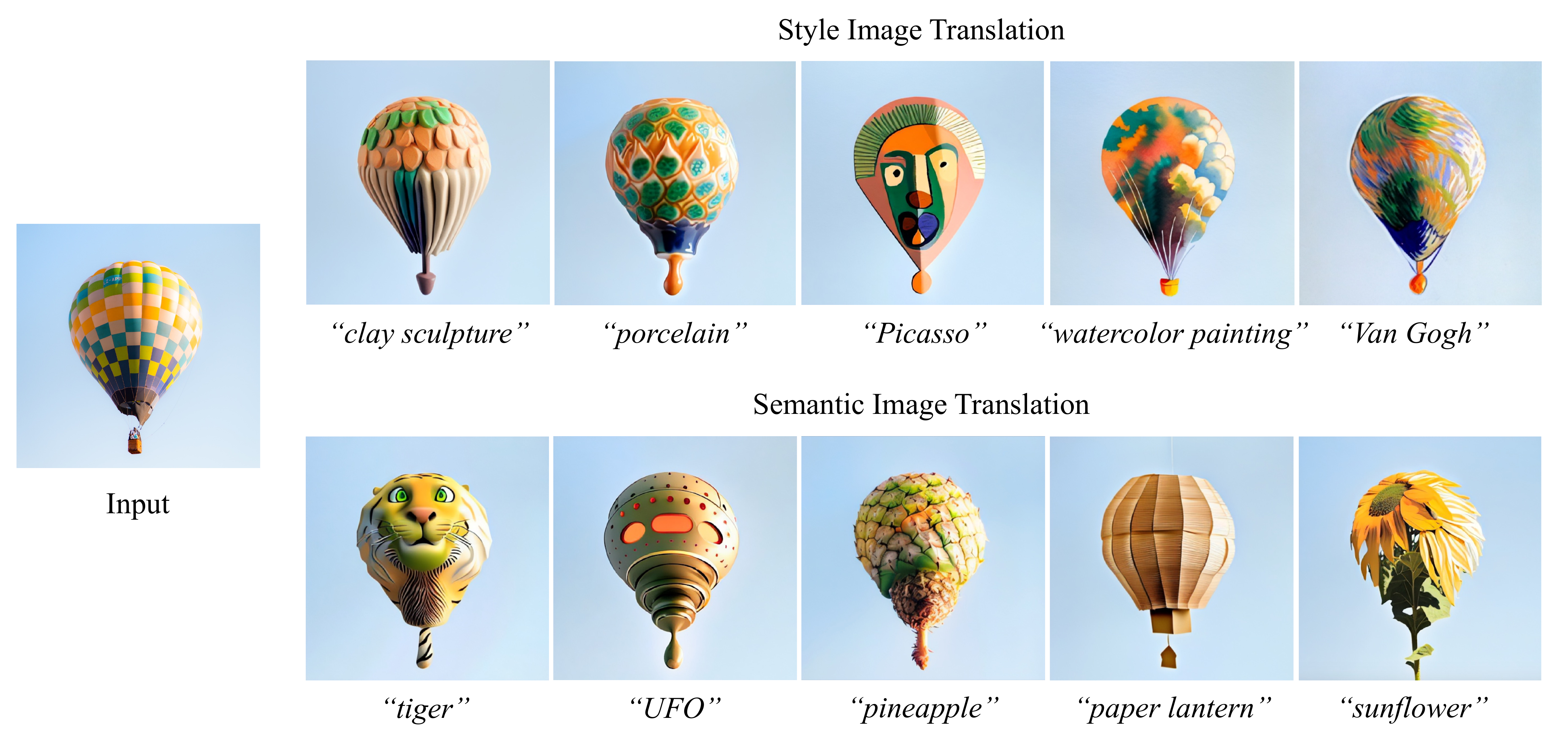}
    \vspace{-0.8cm}
    \captionof{figure}{Style image translation and semantic image translation on real image. Our method can accomplish image translation of arbitrary style and semantics while keeping the layout unchanged. Meanwhile, the inference stage does not require any additional training. Image credit (input images): Unsplash.
    }
    \label{fig:teaser}
\end{center}
}]

\begin{abstract}
Diffusion models are able to generate photorealistic images in arbitrary scenes. However,  when applying diffusion models to image translation, there exists a trade-off between maintaining spatial structure and high-quality content. Besides, existing methods are mainly based on test-time optimization or fine-tuning model for each input image, which are extremely time-consuming for practical applications. To address these issues, we propose a new approach for flexible image translation by learning a layout-aware image condition together with a text condition. Specifically, our method co-encodes images and text into a new domain during the training phase. In the inference stage, we can choose images/text or both as the conditions for each time step, which gives users more flexible control over layout and content. Experimental comparisons of our method with state-of-the-art methods demonstrate our model performs best in both style image translation and semantic image translation and took the shortest time. 
\end{abstract}

\blfootnote{*Denotes equal contribution}

\section{Introduction}
\label{sec:intro}

Image translation tasks usually convert input images to a specific target domain, and require the generated target domain images and source domain images to be consistent in spatial layout \cite{isola2017image,choi2018stargan}. Most of the traditional image translation algorithms cannot make major changes to the global content, which makes them only able to complete the task of image translation in a limited range. The great success of diffusion models in image generation has enabled arbitrary text-based image translation \cite{ho2020denoising,saharia2022photorealistic,ramesh2022hierarchical,rombach2022high}. For example, the high controllability of the layout can not only inspire designers, but also help improve efficiency. However, the text-to-image mapping is one-to-many and the guidance of the text may conflict with the input image. For example, when we convert a real scene to an anime scene, a large deformation often occurs. The bigger the change to the original image, the more obvious the corresponding style will be. When the common layout of the guided text is quite different from the original layout of the input image, the current diffusion-based methods often cause significant changes to the spatial layout of the original image, even though some of these methods use the information of the input image as initialization before denoising \cite{meng2021sdedit}. This does not meet the requirements for image translation \cite{kwon2022diffusion}. 

Recently, some general image editing methods have been proposed and are able to alleviate the above problems to a certain extent \cite{hertz2022prompt,kawar2022imagic,ruiz2022dreambooth,kwon2022diffusion,valevski2022unitune}. However, some of them require the introduction of clip-guided loss during inference to continuously optimize the generator \cite{kim2022diffusionclip,kwon2022diffusion}, while others require individual training or fine-tuning of the entire generative model for each input image \cite{hertz2022prompt,kawar2022imagic,ruiz2022dreambooth,valevski2022unitune}. This is unfavorable for common users, and also time-consuming. Meanwhile, since most of them are not aimed at image translation tasks, their effect is weak in maintaining the spatial information of the input image.

To address the above issues, we propose a novel diffusion-based method, Disign Booster, aims to better balance the trade-off between text description and the input image layout preservation when driving image translation with arbitrary text. Moreover, the proposed method avoids changing the model weights during inference or performing individual optimizations for each image. In addition, inspired by the 
Stable Diffusion model \cite{rombach2022high}, we also adopt a similar approach to further improve the training speed and inference efficiency of the diffusion model. As a result, our method has an inference speed of seconds.

To achieve the above capabilities, we make two significant changes. Firstly, we introduce an encoder jointly trained with the diffusion model to extract the spatial information from the original input image. It is worth noting that we do not directly concatenate the image condition on the initialization noise of the diffusion model as related works \cite{rombach2022high,lugmayr2022repaint}. Considering the generality of the model, we did not use the semantic segmentation map of the input image or the object detection layout as the input of the model, because the identifiable categories are limited and the effect is difficult to guarantee. At the same time, we observe that simple concatenation causes the model to quickly overfit to the image condition during the training phase, making the text condition difficult to learn. Secondly, our model has two conditions (input image and text) during training, and both conditions have a certain percentage of dropout during the training phase. Benefiting from this, our model can flexibly choose control conditions (only text, only images, or both text and images) at each time step in the denoising stage of the diffusion model, and thus can more flexibly control the final generated results. Considering that the early time step of the diffusion process is generating the layout, and the later is adding details, we uniformly only give the input image as a condition in the early time step. However, it is worth noting that different conditions can be adopted for each step here, which provides users with more flexible and controllable parameter choices for personalizing each image.

To sum up, the major contributions of the paper include:
\begin{itemize}
\item For the first time, we involve both text and image conditions in the training phase. We propose a novel network structure which enables text-guided image translation in the inference phase without any additional training burden.
\item We propose an appropriate length ratio for text and image embeddings, which effectively balances the two's ability to control the effect of model generation.
\item We also propose a flexible inference method: using different conditional inputs at different time steps (for example, if you need to maintain more structure, you can use the input as a condition for more time steps, and vice versa).
\item Experimental results see powerful generalizability of our model. Our method performs well on both style image translation and semantic image translation tasks.
\end{itemize}

\begin{figure*}
\begin{center}

    \centering
    \vspace{-0.7cm}
    \includegraphics[width=1\textwidth]{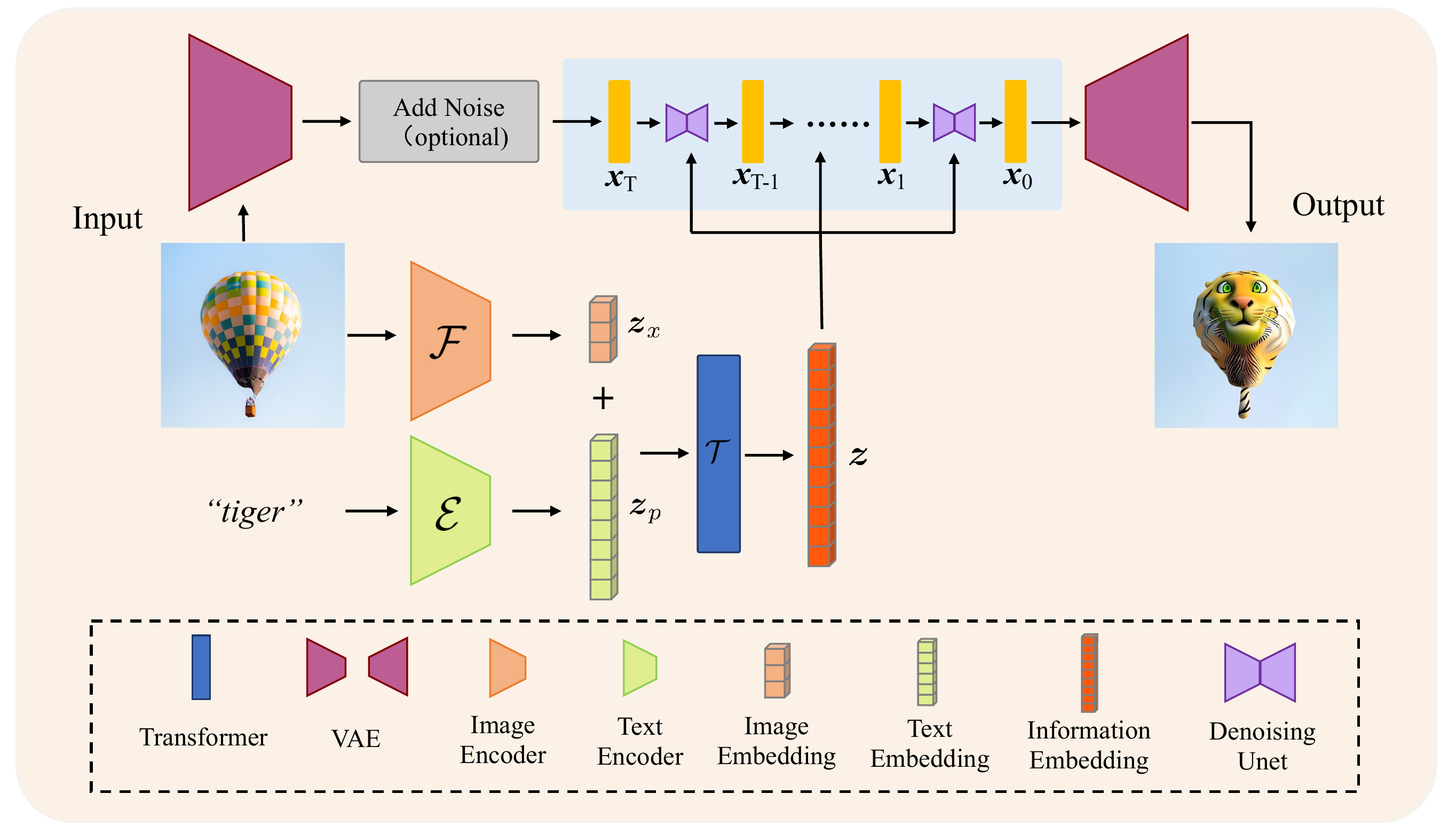}
    \caption{Overview of Design Booster. The structure consists of a frozen text encoder $\bm{\mathcal{E}}:\bm{p} \rightarrow \bm{z}_{p}$ that maps prompt text into text feature, and a trainable image encoder $\bm{\mathcal{F}}:\bar{\bm{x}}_0 \rightarrow \bm{z}_{x}$ which encodes a condition image into image feature. Then the concatenated feature is injected into a transformer $\bm{\mathcal{T}}:\textit{cat}(\bm{z}_{p}, \bm{z}_{x}) \rightarrow \bm{z}$. Based on the joined condition $\bm{z}$, models can be trained and images can be sampled thereby.}. 
    \vspace{-0.5cm}
    \label{overview}
\end{center}
\end{figure*}

\section{Related Work}
\label{sec:related-work}
\subsection{Text-to-Image Synthesis}
As language text naturally describes the intention of humans, a lot of progress has been made in text-to-image synthesis recently due to its significance. The early attempts achieve domain-specific generation by injecting text into generative adversarial nets (GAN) as a condition~\cite{zhang2017stackgan,xu2018attngan,zhu2019dm,li2020manigan}. After that a group of methods~\cite{ramesh2021zero, ding2021cogview, wu2022nuwa, gafni2022make} obtain impressive performance in arbitrary domains based on auto-regressive transformer and VQ-VAE framework~\cite{van2017neural,esser2021taming}. More recently, the diffusion model~\cite{sohl2015deep,song2020score,ho2020denoising} substantially improves the quality and diversity of text-to-image synthesis. Important progress such as DALL-E 2~\cite{ramesh2022hierarchical}, Imagen~\cite{saharia2022photorealistic} and Stable Diffusion~\cite{rombach2022high} are typically trained on large-scale data and can generate photo-realistic images. However, how to effectively leverage the powerful diffusion models for semantic and style manipulation has not been fully explored. Several recent advancements propose different strategies to facilitate diffusion-based image editing. Textual Inversion~\cite{gal2022image} adapts a pretrained diffusion model to specific subject generation by learning a textual token, which can explicitly represents a concept (\ie, content or style) and further recreates the concept through the text-to-image model. DreamBooth~\cite{ruiz2022dreambooth} fine-tunes the diffusion model and a rare token simultaneously to derive the details of a concept. Imagic~\cite{kawar2022imagic} aligns image and text jointly using optimization and fine-tuning strategies. The above methods indeed show excellent manipulation ability meanwhile maintaining visual fidelity. However, lacking the consideration of spatial layout, these methods perform poorly in preserving content structure.


\subsection{Text-Guided Image-to-Image Translation}

Typical methods for image-to-image translation mainly focus on limited domains translation~\cite{isola2017image,zhu2017unpaired,choi2018stargan} or single-shot translation~\cite{cohen2019bidirectional,lin2020tuigan,tumanyan2022splicing,deng2022stytr2}. Text-guided image translation provides the possibility for zero-shot image translation, in which the target content or style are specified by natural language rather than images. To achieve this goal, many methods utilize a multimodal model (\eg, CLIP~\cite{radford2021learning}) as a guidance model. StyleCLIP~\cite{patashnik2021styleclip} proposes several schemes to manipulate images, which modifies the latent vectors of images in StyleGAN~\cite{karras2019style} based on latent optimization and latent encoding. DiffusionCLIP~\cite{kim2022diffusionclip} first obtains the initial noise by a deterministic DDIM process and then fine-tunes the model guided by CLIP loss. DiffuseIT~\cite{kwon2022diffusion} designs a denoising generation process that self-attention keys from ViT model are used to preserve image content, and meanwhile the CLIP loss is applied to guide style. FlexIT~\cite{couairon2022flexit} projects text and image into a same multimodal embedding space, and then performs optimization based on different regularization terms. CLIPstyler~\cite{kwon2022clipstyler} introduces a patch-wise CLIP loss to learn a StyleNet. A different roadmap is translating images based on pretrained text-to-image model. For example, StableDiffusion~\cite{rombach2021highresolution} applies SDEdit~\cite{meng2021sdedit} for image translation, which first adds noise to images within specific steps, and then images are denoised to target domain. The above methods either suffer from unsatisfactory image quality, or struggle to preserve the layout of input images.

\begin{figure*}[htb]
\begin{center}
    \centering
    \vspace{-0.7cm}
    \includegraphics[width=0.9\textwidth]{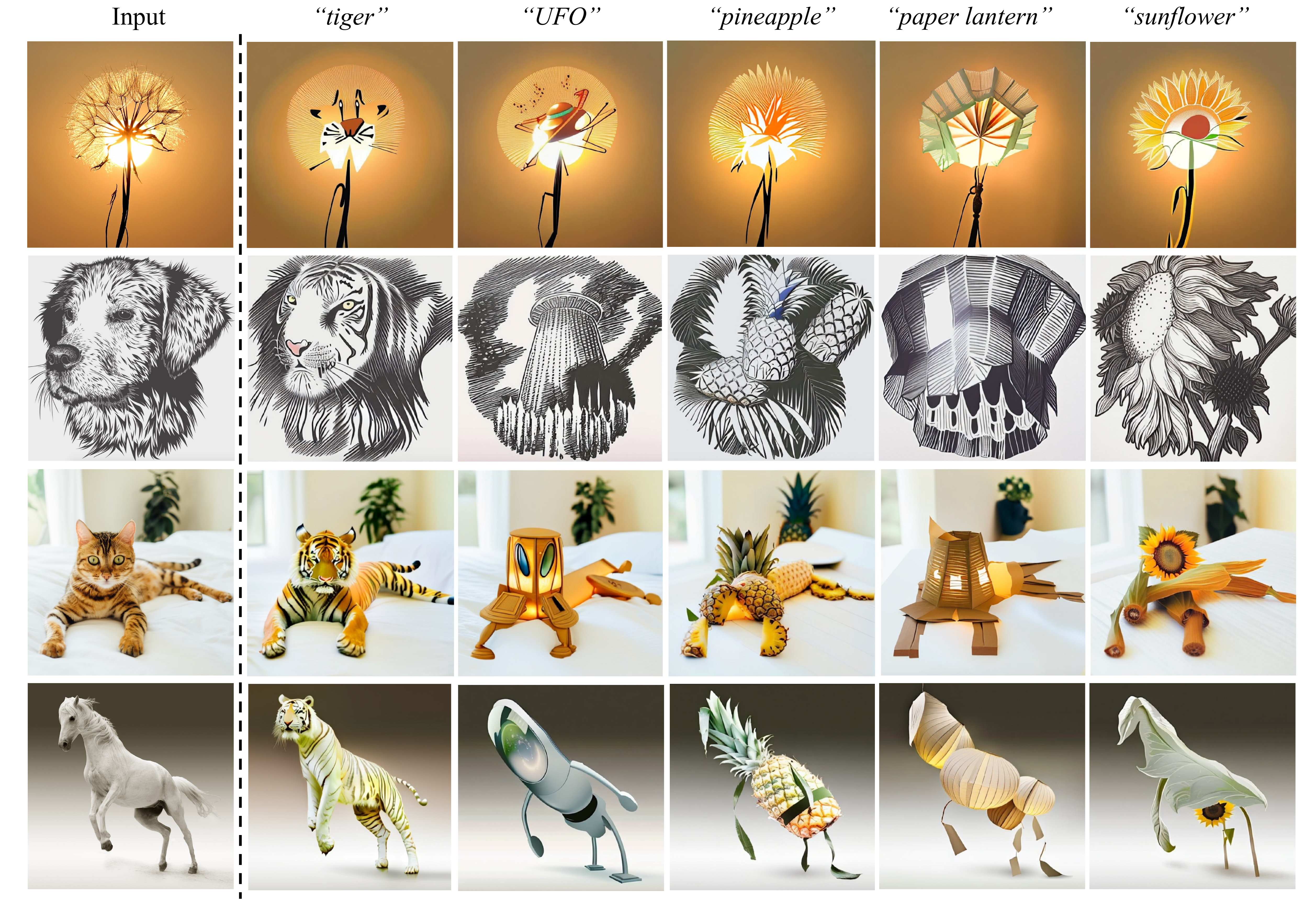}
    \caption{Semantic image translation on real images.}
    \vspace{-0.8cm}
    \label{fig:semantic_2}
\end{center}
\end{figure*}
\section{Proposed Method}

\subsection{Preliminaries}
\label{Preliminaries}

Denoising Diffusion Probabilistic Model (DDPM)~\cite{ho2020denoising} is a commonly used diffusion model, which consists of a forward diffusion process and a reverse diffusion process. Given an image $\bm{x}_0$ in the training set, each step in the forward process gradually adds noise through a Markov chain, \ie, $q(\bm{x}_t|\bm{x}_{t-1})=\mathcal{N}(\bm{x}_t;\sqrt{1-\beta_t}\bm{x}_{t-1},\beta_t\bm{I})$, where $\beta_t \in (0, 1)$ is the variance of Gaussian noise. Correspondingly, we have latent vectors $\bm{x}_1, \ldots, \bm{x}_T$ and the final state $\bm{x}_T \sim \mathcal{N}(0, \bm{I})$. Note that $\bm{x}_t$ can be obtained directly from $\bm{x}_0$ as follows:
\begin{equation}
\label{eq:forward}
\bm{x}_t = \sqrt{\bar{\alpha}_t}\bm{x}_0 + \sqrt{1-\bar{\alpha}_t}\bm{\epsilon},
\end{equation}
where $\alpha_t=1-\beta_t$, $\bar{\alpha}_t=\prod_{s=1}^t\alpha_s$ and $\bm{\epsilon} \sim \mathcal{N}(0,\bm{I})$.
The reverse process is also a Gaussian Markov chain:
\begin{equation}
\label{eq:reverse}
p_\theta(\bm{x}_{t-1}|\bm{x}_{t})=\mathcal{N}(\bm{x}_{t-1};\bm{\mu}_\theta(\bm{x}_t, t), \bm{\Sigma}_\theta(\bm{x}_t, t)\bm{I}).
\end{equation}


In DDPM, $\bm{\mu}_\theta$ is indirectly parameterized by $\bm{\epsilon}_\theta$, \ie, $\bm{\mu}_\theta(\bm{x}_t, t)=\frac{1}{\sqrt{\alpha_t}}\big(\bm{x}_t-\frac{\beta_t}{\sqrt{1-\bar{\alpha}_t}}\bm{\epsilon}_\theta(\bm{x}_t, t)\big)$. $\bm{\Sigma}_\theta$ can be set to time dependent constants~\cite{ho2020denoising} or learnable variables~\cite{nichol2021improved}. The conditional training of the parameterized neural network uses the following objective:
\begin{equation}
\label{train objective}
L_{\text{simple}} = \mathbb{E}_{\bm{x}_0, \bm{\epsilon}, \bm{c}} \Big[\|\bm{\epsilon} - \bm{\epsilon}_\theta(\sqrt{\bar{\alpha}_t}\bm{x}_0 + \sqrt{1 - \bar{\alpha}_t}\bm{\epsilon}, \bm{c})\|^2 \Big],
\end{equation}
where $\bm{c}$ is the condition, \eg, class label, text description, \etc.

Classifier-free guidance~\cite{ho2021classifier} is an effective technique which enables us to elegantly inject conditions into diffusion generation process. Specifically, $\bm{\epsilon}_\theta(\bm{x}_t,\bm{c})$ is replaced by a corresponding guided version:
\begin{equation}  
\tilde{\bm{\epsilon}}_\theta(\bm{x}_t, \bm{c}) = (1+s) \cdot {\bm{\epsilon}_\theta(\bm{x}_t,\bm{c})} - s \cdot {\bm{\epsilon}_\theta(\bm{x}_t)}, 
\end{equation} 
where $\bm{\epsilon}_\theta(\bm{x}_t)$ and $\bm{\epsilon}_\theta(\bm{x}_t,\bm{c})$ are unconditional and conditional $\bm{\epsilon}$-predictions. $s$ is guidance scale.

A feasible way to achieve image translation based on off-the-shelf diffusion models is SDEdit~\cite{meng2021sdedit}, which first encodes an initial condition image $\bar{\bm{x}}_0$ to $\bar{\bm{x}}_k$ accordding to Equation~\ref{eq:forward}, where $k$ is a hyperparameter. Then the denoising process starts at $\bar{\bm{x}}_k$ and the target distribution $p_\theta(\bm{x}_0)$ is obtained using Equation~\ref{eq:reverse}.

\begin{figure*}[t]
\begin{center}
    \centering
    \vspace{-0.5cm}
    \includegraphics[width=0.82\textwidth]{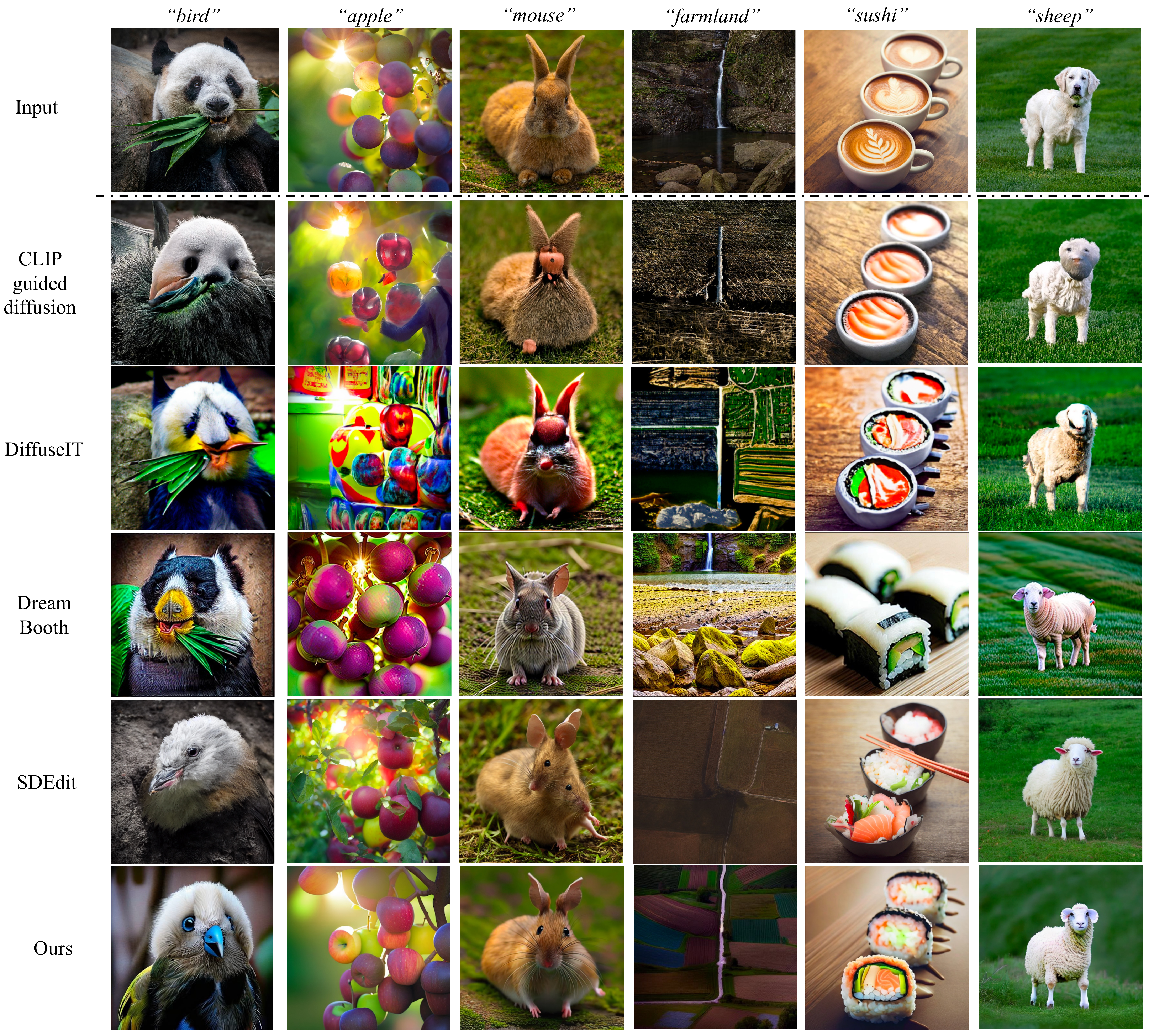}
    \caption{The qualitative comparison on semantic image translation.  Design Booster can give consideration to both the generation quality and spatial preservation.}
    \vspace{-0.7cm}
    \label{fig:semantic_compare}
\end{center}
\end{figure*}

\subsection{Multi-Condition Network}

In order to obtain the target distribution $p_\theta(\bm{x}_0)$ conditioned together on prompt text $\bm{p}$ and initial image $\bar{\bm{x}}_0$, our first task is to effectively fuse the features in text domain and image domain. To achieve this goal, a plausible way is to concatenate the image condition with the diffusion noise directly. However, we empirically find this leads to serious overfitting of the model on image condition, which weakens the guiding role of text condition. To solve the above problem, we propose a novel network. The overview is depicted 
in Figure \ref{overview}.

Generally, a pretrained text encoder play a vital role (\eg, CLIP\cite{radford2021learning}, T5\cite{raffel2020exploring}) to text-to-image model, which can capture the complexity of arbitrary natural language. In this paper, CLIP is used as a frozen text encoder $\bm{\mathcal{E}}$. Given the text condition $\bm{p}$, the text feature is first obtained, \ie, $\bm{z}_{p}=\bm{\mathcal{E}}(\bm{p})$. Then we define an image encoder $\bm{\mathcal{F}}$, and thus the corresponding image feature is $\bm{z}_{x}=\bm{\mathcal{F}}(\bar{\bm{x}}_0)$. The image encoder $\bm{\mathcal{F}}$ aims to summarize the spatial information of image condition $\bar{\bm{x}}_0$ into a descriptive vector  $\bm{z}_{x}$, which stays in the same embedding domain as the text feature $\bm{z}_{p}$. Note that $\bm{\mathcal{F}}$ can be implemented by any neural networks that can process the images. A simple encoder that based on CNN is employed in order to enable fast training and inference, whose detailed architecture can be found in the Appendix. 
In addition, the concatenated embedding $(\bm{z}_{p}, \bm{z}_{x})$ is injected into a transformer modules $\bm{\mathcal{T}}$, which further reduces the gap between the text and image domains:
$\bm{z}=\bm{\mathcal{T}}(\textit{cat}(\bm{z}_{p}, \bm{z}_{x}))$, where $\bm{z}$ is the final multi-condition embedding. Then the conditional reverse process is defined as: $p_\theta(\bm{x}_{t-1}|\bm{x}_{t},\bm{z})=\mathcal{N}(\bm{x}_{t-1};\bm{\mu}_\theta(\bm{x}_t, t, \bm{z}), \bm{\Sigma}_\theta(\bm{x}_t, t)\bm{I})$, and the training objective in Equation \ref{train objective} is redefined as follows:
\begin{equation}
\label{redefined train objective}
    L_{\text{simple}} = \mathbb{E}_{\bm{x}_0, \bm{\epsilon}, \bm{z}} \Big[\|\bm{\epsilon} - \bm{\epsilon}_\theta(\sqrt{\bar{\alpha}_t}\bm{x}_0 + \sqrt{1 - \bar{\alpha}_t}\bm{\epsilon}, \bm{z})\|^2 \Big].
\end{equation}

As the conditional embedding $\bm{z}$ contains information of text and image, the entire model will be optimized in the direction driven by both the text and image conditions. Therefore, the generated results can preserve the layout information from image condition, which cannot be described by text condition.

\begin{figure*}[htb]
\begin{center}
    \centering
    \vspace{-0.5cm}
    \includegraphics[width=0.82\textwidth]{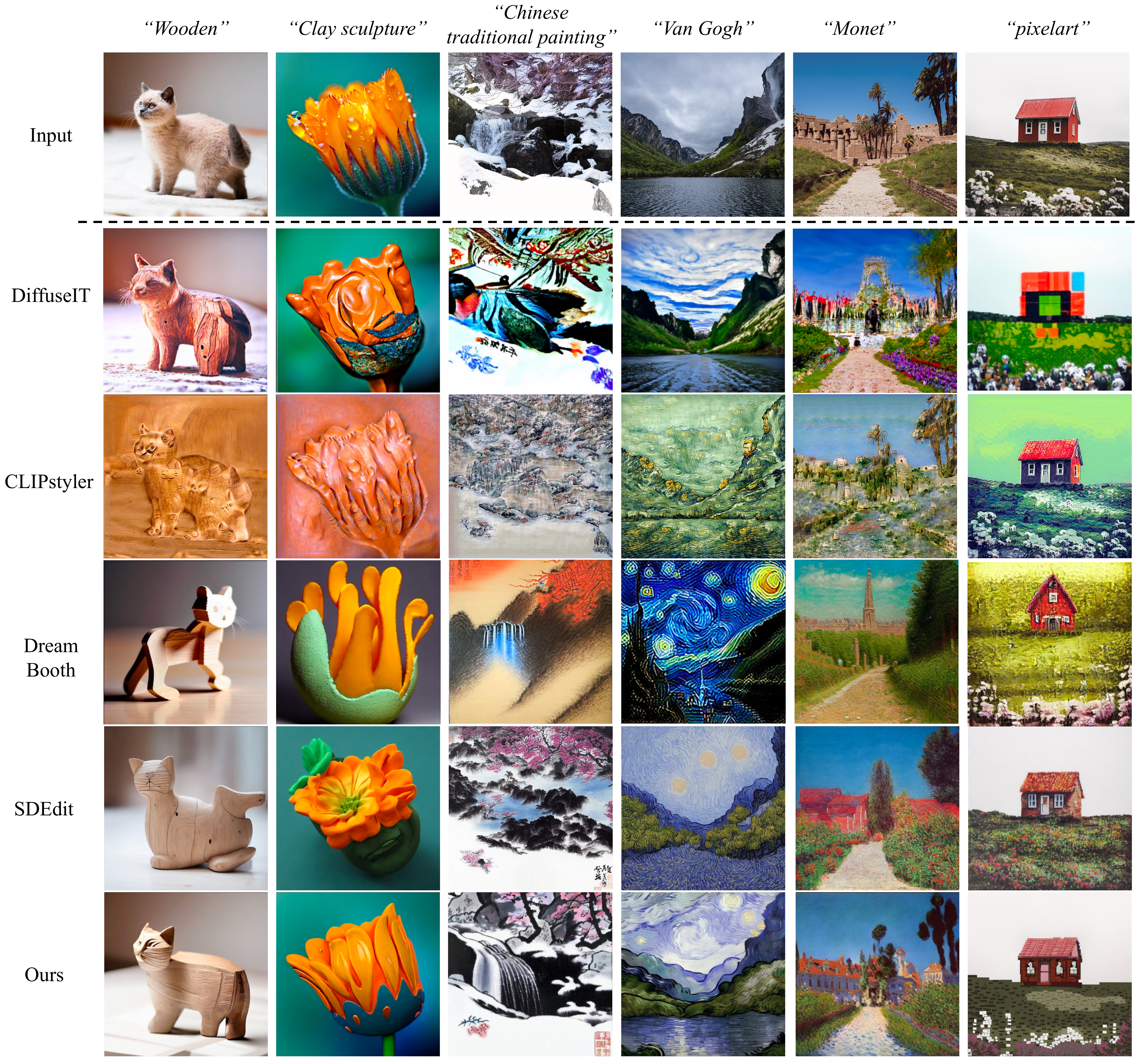}
    \caption{The qualitative comparison on style image translation. Design Booster can generate high-quality stylized results  with the spatial layout preservation of the original image.}
    \vspace{-0.5cm}
    \label{fig:style_compare}
\end{center}
\end{figure*}

\subsection{Flexible Sampling with Multi-condition}
\label{Flexible sampling with multi-condition}

With the condition $\bm{z}$, our model can generate image associated with multi-condition. However, in some cases, it is expected to flexibly control the influence of image or text conditions, \eg, flexibly enlarging or reducing the impact of image condition. Therefore, we propose a mechanism, combined with classifier-free technique~\cite{ho2022classifier}, to solve the above problem.

During training stage, the text condition and image conditions dynamically drop out with probabilities of $d_{p}, d_{x}$ (\eg, dropping the text condition with probability $d_{p}$). During sampling stage, the denoising process can be denoted as Equation~\ref{eq:reverse} with $T$ steps. For each step, the desired conditions can be injected as necessary. To appropriately set the conditions for each step, we design a simple yet effective switch strategy for controlling conditions:
\begin{equation}
    \bm{S} = \{ \bm{z}_{T},\bm{z}_{T-1}, \cdots,\bm{z}_{\sigma},\cdots,\bm{z}_{2},\bm{z}_{1}\},
\end{equation}
\begin{equation}
     \bm{z}_{t} = 
     \begin{cases}
     \bm{\mathcal{T}}(\textit{cat}(\bm{z}_{p}, \bm{z}_{x})), \quad\text{if} \quad t >= \sigma, \\
     \bm{\mathcal{T}}(\textit{cat}(\bm{z}_{p} , \bm{z}_{\phi})),  \quad \text{otherwise},
     \end{cases}
\end{equation}
where $t$ is the diffusion step. $\sigma$ is a hyperparameter denoting the switch step. $\bm{z}_{\phi}$ is a empty image embedding. This strategy is based on the fact that the denoising process always first generates the spatial layout of the image, and then determines the content and refines the details~\cite{choi2022perception}. Therefore, the image condition is added at the beginning of the denoising process. The more steps the image conditions are added, the closer appearance the synthesized images are compared with the condition image.
The steps that image conditions are added is determined by $\sigma$, and we discuss the effect of $\sigma$ on the generated results in Section \ref{Ablation}. With the proposed  multi-condition strategy, the classifier-free guidance is redefined as follows:
\begin{equation}  
\tilde{\epsilon}_\theta(\bm{x}_t,\bm{z}_{t}) = (1+s) \cdot {\epsilon_\theta(\bm{x}_t,\bm{z}_{t})} - s \cdot {\epsilon_\theta(\bm{x}_t)},   
\end{equation}
where $\bm{z}_{t} \in \bm{S}$. Therefore, we can control the final generated results through the multi-condition $\bm{z}_{t}$ within various strategy $\bm{S}$.

\section{Experiments}
The training details is described in Section \ref{Training details}. Some properties of our method are assessed in Section \ref{Conditions affect}. The two main downstream tasks, style image translation and semantic image translation, are described in Section \ref{style image translation} and Section \ref{semantic image translation}. Since there is no ground truth, it is a bit far-fetched to evaluate our method with quantitative metrics. In the Appendix, indicators of structural similarity are tentatively used to make a quantitative evaluation.

\subsection{Training Details}
\label{Training details}

We deploy our method on Stable Diffusion\footnote{https://github.com/CompVis/stable-diffusion}~\cite{rombach2022high}. The key configurations follow the official setting, including an autoencoder with a downsampling factor 8, a UNet with 860M parameters and a CLIP~\cite{radford2021learning} text encoder implemented by ViT-L/14. The image encoder $\bm{\mathcal{F}}$ is a lightweight network based on conlutional layers, which is detailed in the Appendix. The model is partially initilized from pretrained Stable Diffusion model to inherit the knowledge of text-to-image model. The whole model is trained on $512 \times 512 $ images of Laion dataset \cite{schuhmann2021laion}, with text and image dropout probabilities of $d_p=0.1$ and $d_x=0.6$. The training batch size is set to 1024.

\subsection{Analysis of Conditions}
\label{Conditions affect}
\begin{figure}[h]
\begin{center}
    \includegraphics[width=0.45\textwidth]{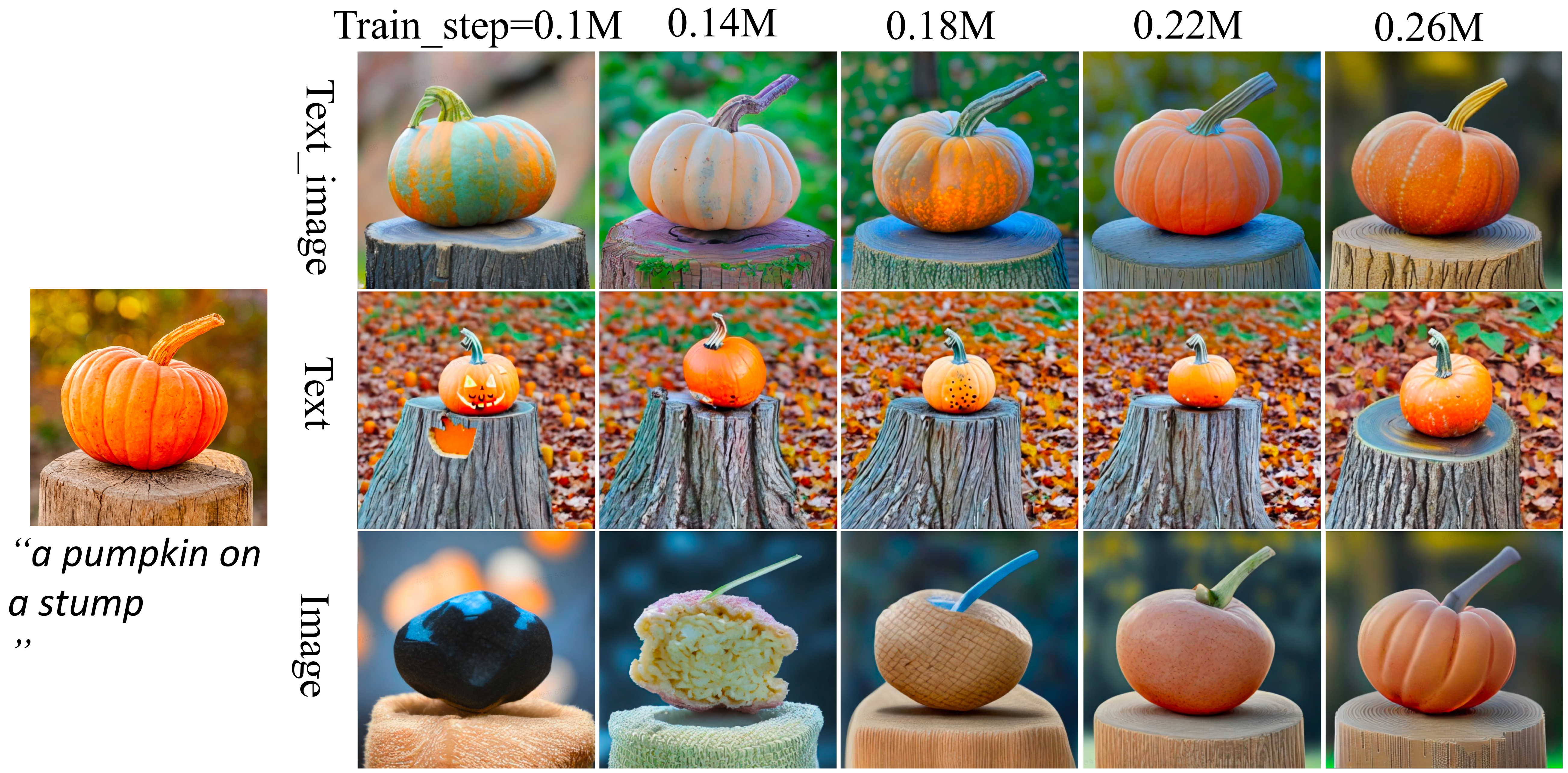}
    \caption{Here are three types of conditions (text and image, only text and only image). Design Booster can learn the features of the image condition continuously.}
    \vspace{-1.0cm}
    \label{fig:con11}
\end{center}
\end{figure}

Each time step of our model can flexibly choose three conditions: 1) text only ($\sigma = 1$); 2) image only ($\sigma = T + 1$); 3) text and image. In order to explore the influence of these three conditions on the generated results, we visualize the generated results with only one condition at all time steps under different training durations in Figure \ref{fig:con11}.

It is shown that when only the image condition is used, the results will be more and more similar to the spatial layout of the input image, even the shape of a pumpkin is generated. Here, the incomplete reconstruction of details is exactly what we want to see, since a perfect reconstruction would result in reduced ability for text-guided detail editing. When we want to greatly modify the color information of the input, we can choose the early training weights, otherwise it is suitable for the longer training model. In this paper, for fair comparison, all results are generated with the model trained for 0.26M iterations. When only the text condition is used, the results are detailed, but the spatial layout is not consistent with the input image. It's worth noting that, with text-only condition, we tested with the same random seed with DDIM~\cite{song2020denoising} and the generated results during training are very similar. So the preloaded text-to-image knowledge is not weakened. Once both conditions are injected, the result is almost the same as input image. 

\subsection{Optional SDEdit}
SDEdit ~\cite{meng2021sdedit} is a simple method for style image translation and semantic image translation, which is presented in Section~\ref{Preliminaries}. The $\textit{strength}$ in SDEdit is used to balance how much the original image is preserved and how much the image is edited. On the one hand, the preservation of the original image information is not enough when the translation is strong. On the other hand, if the original image information is retained too much, the translation effect will not be obvious. Different from SDEdit, our method is more flexible which can inject the image condition in any steps during the denoising process. SDEdit is treated as an optional mechanism in our method. More effects of this mechanism will be discussed in Section \ref{Ablation}.


\subsection{Style Image Translation}
\label{style image translation}
In order to evaluate the performance of our model for style image translation, we conduct the comparisons with the state-of-art methods. DiffuseIT\footnote{https://github.com/anon294384/diffuseit}~\cite{kwon2022diffusion}, CLIPstyler\footnote{https://github.com/cyclomon/CLIPstyler}~\cite{kwon2022clipstyler}, DreamBooth~\cite{ruiz2022dreambooth} and SDEdit~\cite{meng2021sdedit} are selected as the baseline methods. We reference the open source codes for the first two methods, DiffuseIT and CLIPstyler.

\begin{figure}[htb]
\begin{center}
    \centering
    \vspace{-0.5cm}
    \includegraphics[width=0.5\textwidth]{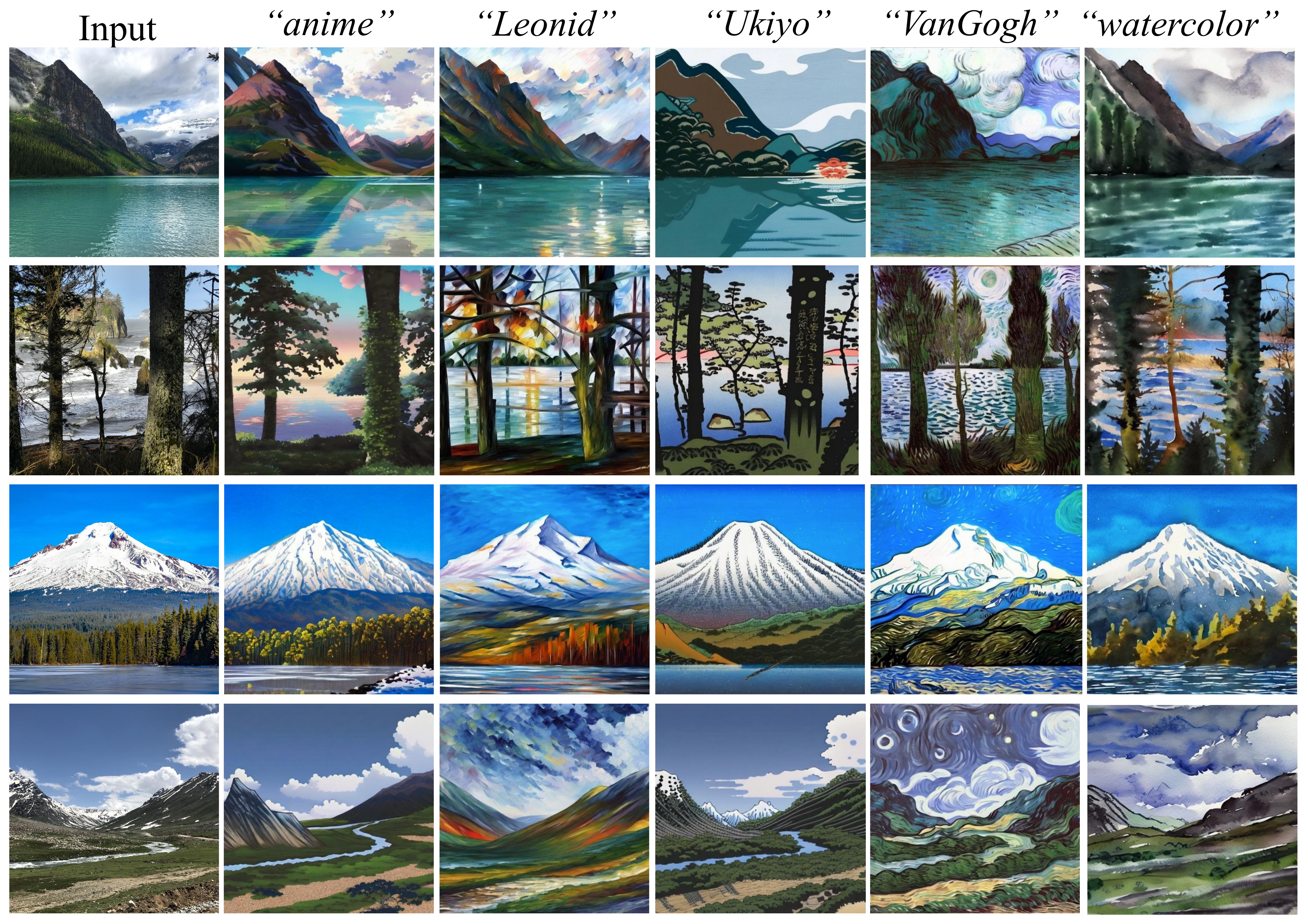}
    \caption{Convert real images to painting style.}
    \vspace{-0.3cm}
    \label{fig:style}
\end{center}
\end{figure}

\begin{figure}[htb]
\begin{center}
    \centering
    \vspace{-0.5cm}
    \includegraphics[width=0.5\textwidth]{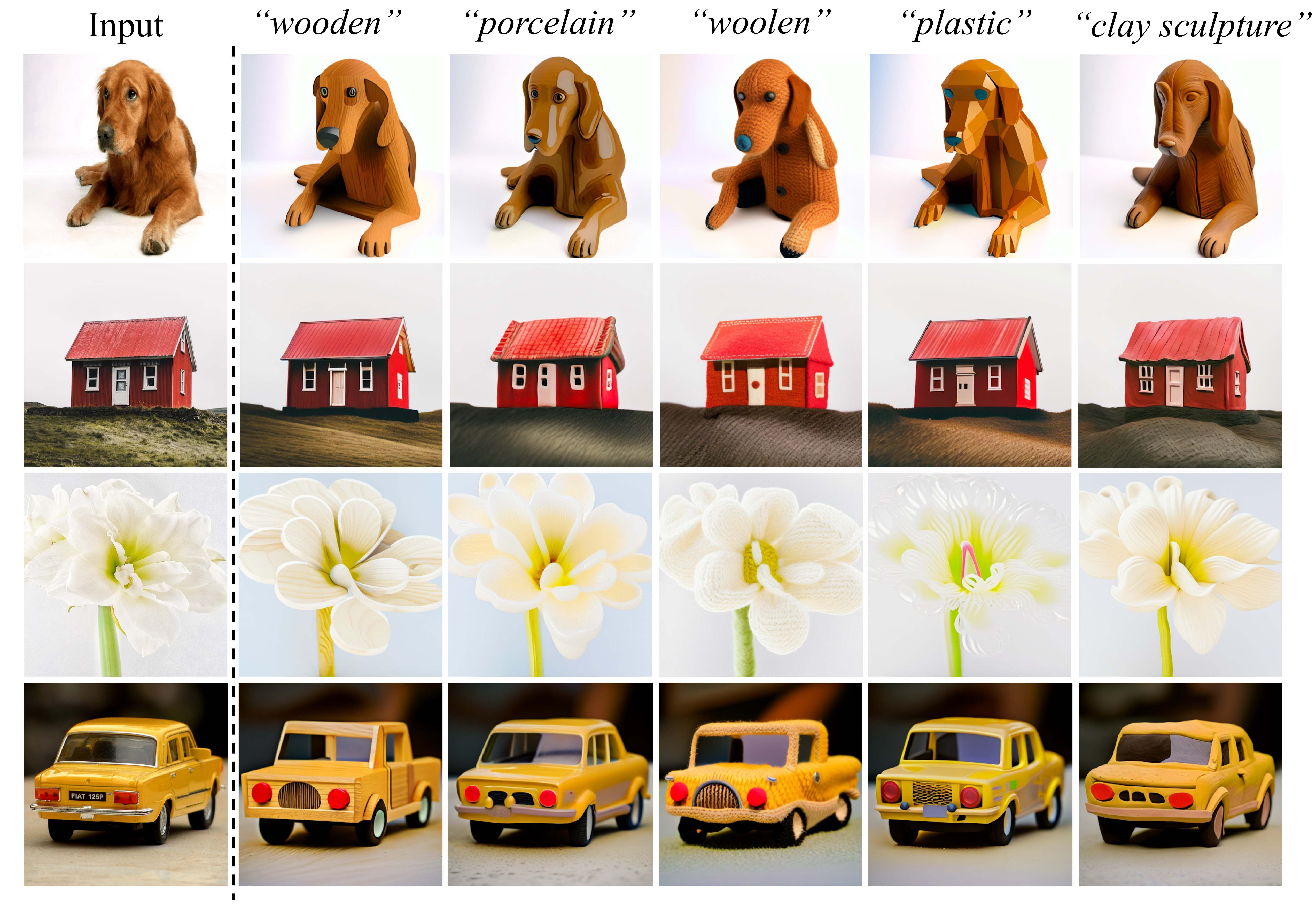}
    \caption{Convert real images to styles of different materials.}
    \vspace{-0.5cm}
    \label{fig:style_texture}
\end{center}
\end{figure}

For fairness, DreamBooth and  SDEdit are also implemented on pretrained Stable Diffusion. Experiments are carried out not only on common styles (\eg, Van Gogh ), but also on the style transfer of materials, which are shown in Figure \ref{fig:style_compare}. The stylization quality of DreamBooth and SDEdit are the closest to our method, but neither of them can effectively preserve the spatial layout of the original image. More examples about the style transfer are given in Figure~\ref{fig:style} and Figure~\ref{fig:style_texture}.

\subsection{Semantic Image Translation}
\label{semantic image translation}




Since Design Booster can been applied on semantic translation easily, we further explore the capability of our method on this task. The comparison experiments are conducted with CGD~\cite{cgd}, DiffuseIT~\cite{kwon2022diffusion}, DreamBooth~\cite{ruiz2022dreambooth} and SDEdit~\cite{meng2021sdedit}. The configures of the comparison methods are the same as the above Section~\ref{style image translation}.

Figure \ref{fig:semantic_compare} shows the generation quality of different methods. We can see that DreamBooth and SDEit can generate the same high quality results, but the spatial layout is deformed severely. The other two methods, CGD and DIffusionIT, have exactly opposite properties. Design Booster can preserve the spatial layout of the original image and gernerate high-quality results in line with text guidance. More results are shown in Figure \ref{fig:semantic_2}.

\subsection{Ablation}
\label{Ablation}
\begin{figure}[h]
\begin{center}
    \centering
    \vspace{-0.5cm}
    \includegraphics[width=0.5\textwidth]{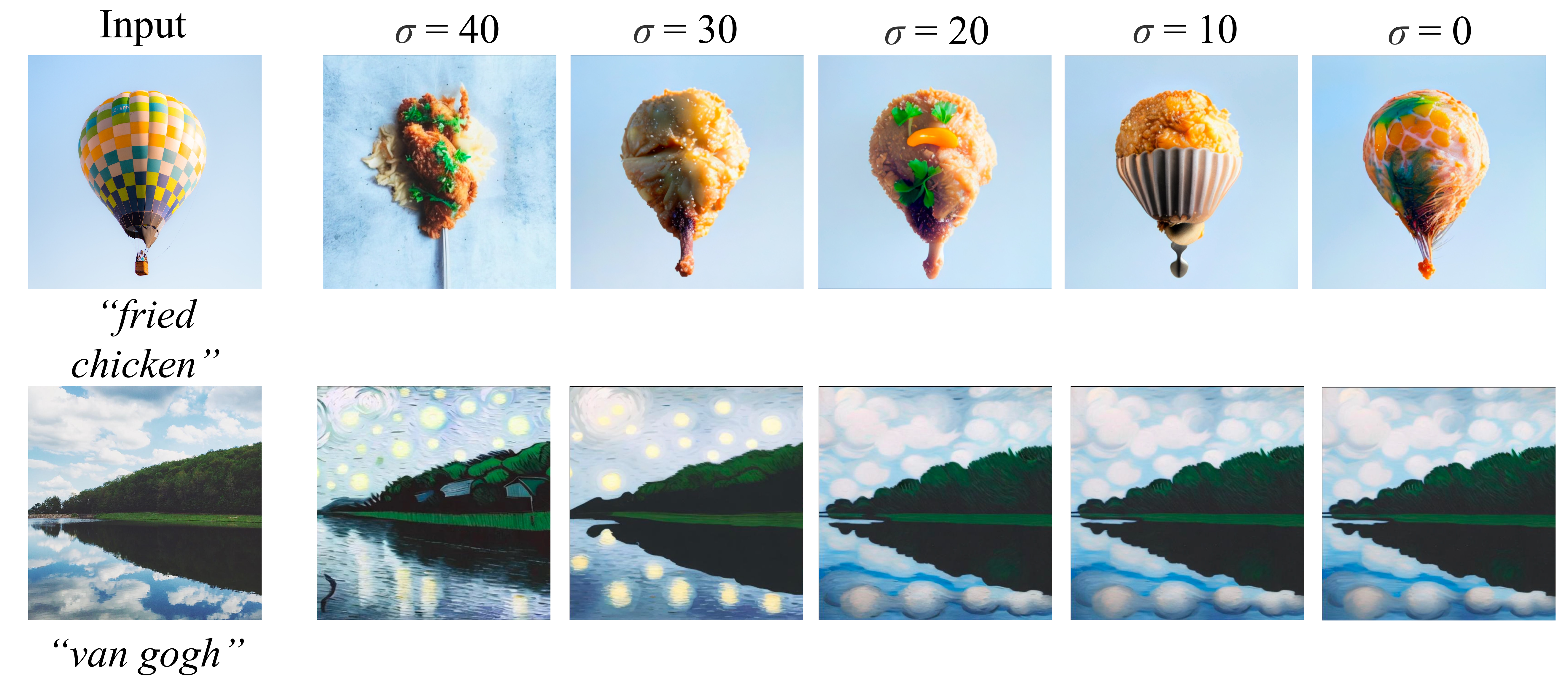}
    \caption{This shows the effect of different $\sigma$ on the generated images. The smaller $\sigma$, the more image information will be added, and the result will be closer to the input image.}
    \vspace{-0.5cm}
    \label{fig:switch_t}
\end{center}
\end{figure}

\paragraph{Analysis of $\sigma$.}
The conditions of the sampling process can be arbitrary in our method, but this can lead to complex permutation of conditions. So a simple and effective strategy, defined in Section~\ref{Flexible sampling with multi-condition}, is proposed by controlling when to inject image conditions, where $\sigma$ is the only parameter. In order to explore the effect of injecting image conditions at different times, the generated images are shown in Figure \ref{fig:switch_t}, with DDIM~\cite{song2020denoising} 50 steps.

\begin{figure}[h]
\begin{center}
    \centering
    \includegraphics[width=0.5\textwidth]{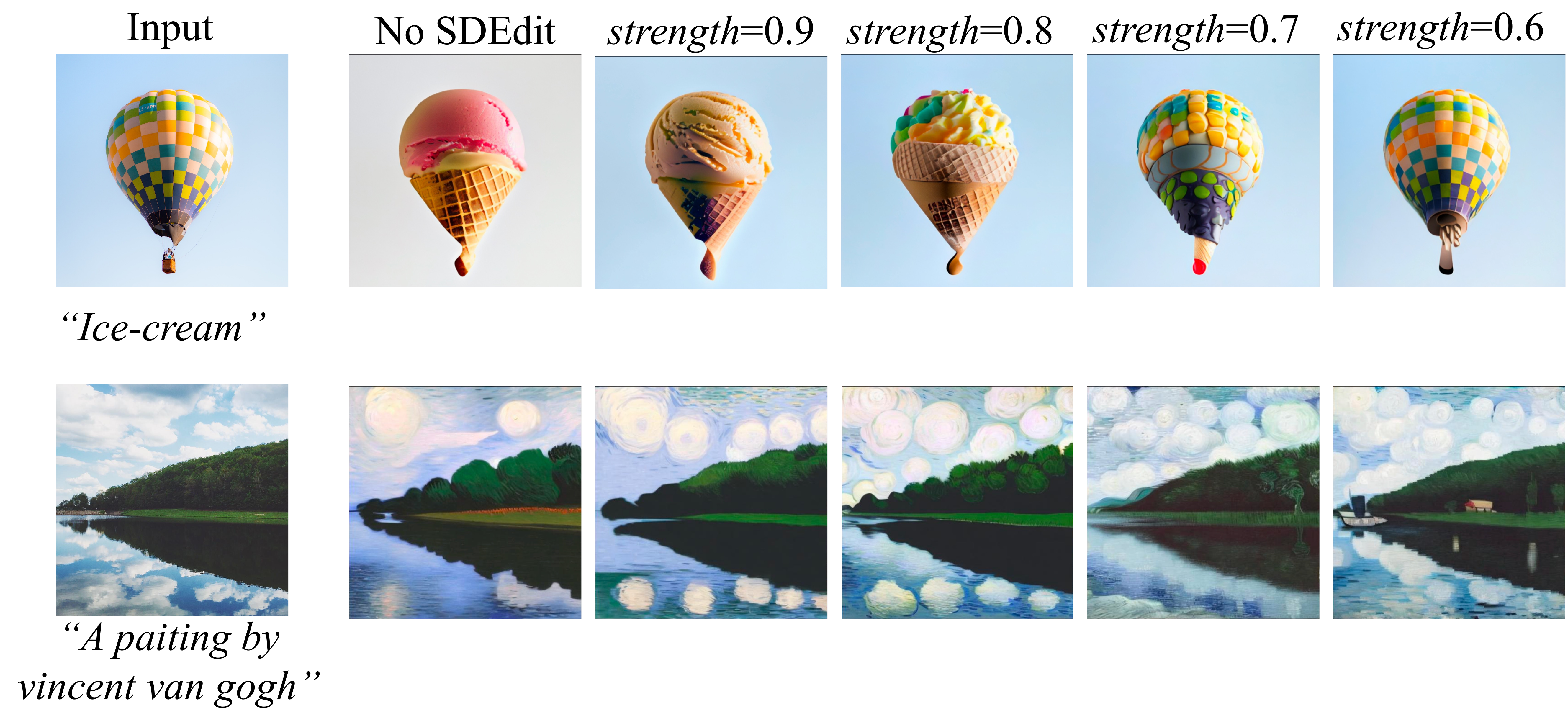}
    \caption{SDEdit is optional in our method. An appropriate \textit{strength} will further improve the correlation with the input image without affecting the generated results}
    \vspace{-0.5cm}
    \label{fig:SDE_strength}
\end{center}
\end{figure}

The generated images in Figure \ref{fig:switch_t} indicate that injecting image conditions will make the generated results have similar appearance as the condition images, both in style image translation and semantic image translation. However, if too many image conditions are added (setting $\sigma$ to a small value), the generated results will show many details of the original image and weaken the guiding role of the text. For example, when we want to turn a hot air balloon into fried chicken and set  $\sigma$ to 0, the result will generate with a lot of details from the  hot air balloons instead of real fried chicken. Therefore, considering that in the semantic image translation task, the text is more likely to destroy the original layout of the image, and thus we choose a smaller $\sigma$. In this paper, $\sigma$ is set to 25 for semantic image translation and 35 for style image translation. In practical applications, this parameter can be changed more flexibly, and even the image and text  conditions can be used at intervals.
\paragraph{Strength in SDEdit.}
The text condition and image condition can be flexible adjusted by $\sigma$. In order to further enhance the correlation between the generated results and the image conditions, our method can be combined with SDEdit.  So it is necessary to set a appropriate $\textit{strength}$. The effect of different $\textit{strength}$ on the generated results is shown in Figure~\ref{fig:SDE_strength}. 

It demonstrates that the results are good enough without SDEdit, after combining, more details can be enhanced to be closer to the condition image. But if $\textit{strength}$ is too small, it will significantly weaken the translation ability. Therefore, our method can be combined with SDEdit ($\textit{strength}$ is set to 0.8)  optionally.

\section{Conclusion}
We propose a diffusion model for image translation, named Design Booster. Design Booster is able to arbitrarily edit the style and semantics of images while preserving the input image layout as much as possible. At the same time, the proposed method does not require any additional training during inference. The comparison with existing methods verifies that our method has significant advantages in both effect and time performance.

Nevertheless, a limitation of our model is that our model has a slightly weaker ability to change color under strong layout-preserving parameters. It is recommended to use input image as a condition in fewer time steps when large changes in color are required.

\section{Appendix}
\subsection{User Study}

We conduct two user study on style and semantic image  translation. For each task , three questions of different dimensions are designed to evaluate the method: 1) Do the results conform to the text condition (translation) 2) Do the results have the same spatial layout as the input image. 3) Are the results are of high quality and realistic. 100 images are generated using  5 text conditions with all methods. We selected 50 images for every user and ask the above three questions. For each question, there are 3 options for each image, UNLIKELY(1 point), NORMAL(2 points) and LIKELY(3 points). And 20 users between the ages of 20 and 50 are recruited. Average scores are shown in Table \ref{tab:user_study_1} and Table \ref{tab:user_study_2} ,compared with other methods. 
\begin{table}[h]
    \centering
    \begin{tabular}{c|ccc|c}
    
        \toprule
        \textbf{Method} &  Translation & Layout& Quality&Avg.$\uparrow$\\
        \hline
        DiffuseIT \cite{kwon2022diffusion}& 1.35 & 2.57 & 1.43&1.78\\
        CLIPstyler \cite{kwon2022clipstyler} & 1.82 & 2.63 & 1.52&1.99\\
        Dream Booth \cite{ruiz2022dreambooth} & 2.27 & 1.57 & 2.28&2.04\\
        SDEdit \cite{meng2021sdedit}& 2.32 & 1.97 & 2.21&2.17\\
        Ours & 2.33 & 2.41 &2.29&\textbf{2.34}\\
        \bottomrule
    \end{tabular}
    \caption{Users study  on style image translation.}
    \label{tab:user_study_1}
\end{table}
\begin{table}[h]
    \centering
    \begin{tabular}{c|ccc|c}
        \toprule
        \textbf{Method} &  Translation & Layout& Quality&Avg.$\uparrow$\\
        \hline
        CGD \cite{cgd}& 1.39 & 2.47 & 1.58 &1.81\\
        DiffuseIT \cite{kwon2022diffusion}& 1.48 & 2.50 &1.76&1.91\\
        Dream Booth \cite{ruiz2022dreambooth} & 2.33 & 1.54 &2.26&2.04\\
        SDEdit \cite{meng2021sdedit} & 2.49 & 2.02 &2.36& 2.29\\
        Ours &  2.47& 2.43 & 2.37&\textbf{2.42}\\
        \bottomrule
    \end{tabular}
    \caption{Users study  on semantic image translation.}
    \label{tab:user_study_2}
\end{table}

Our method has the highest average score for the three questions. It demonstrates  that our method can accomplish image translation of arbitrary style and semantics while keeping the layout unchanged.



\subsection{Knowledge of Pretrained Model }
\begin{figure}[h]
\begin{center}
    \includegraphics[width=0.5\textwidth]{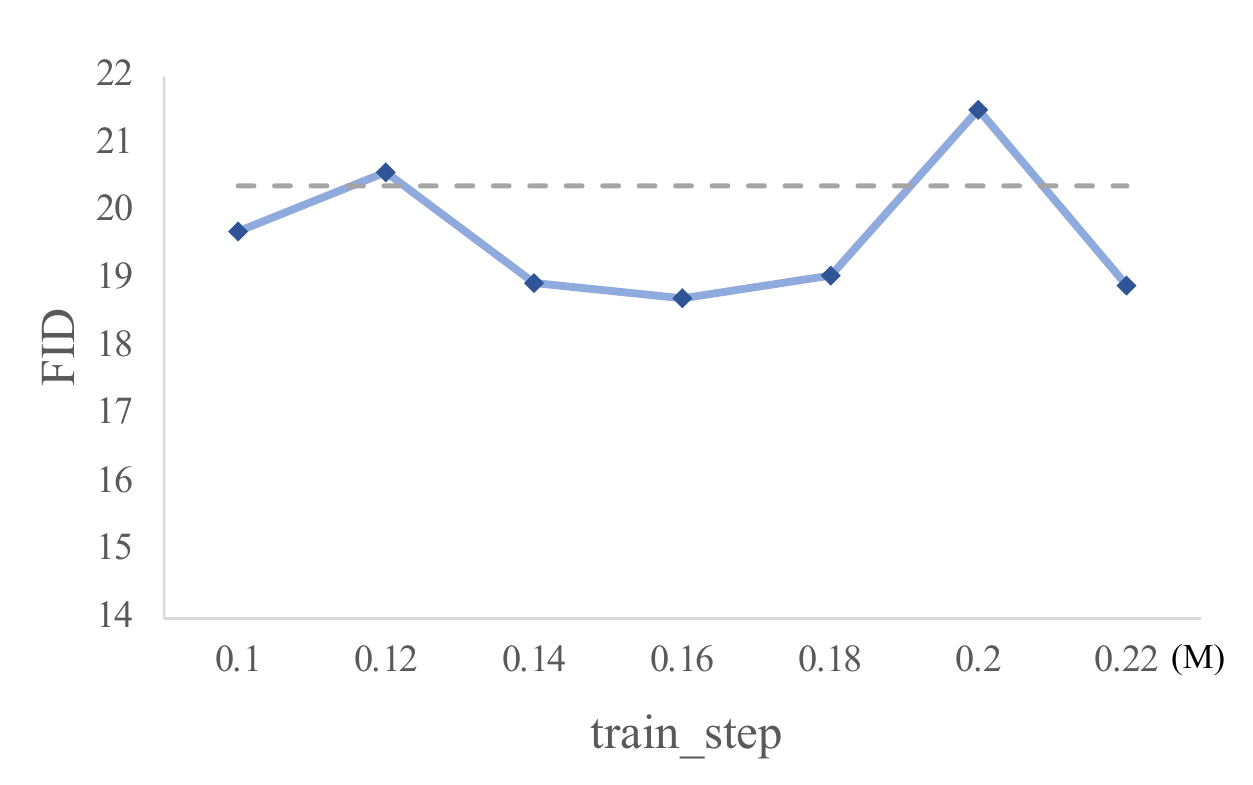}
    \caption{This figure shows the FID of our model during training. And the dashed line indicates the FID of the preloaded model}
    \label{fig:FID}
\end{center}
\end{figure}

To save computing resources, we preload the weight from text-to-image model. When the image condition is null, our model can also be viewed as a text-to-image model. Therefore, a noticeable question is whether our method will weaken the generation ability of the pretrained model. To quantify this problem, we measured the FID-30K of the trained pretrained model on COCO \cite{lin2014microsoft} ,as well as our model during training. The result is shown in Figure \ref{fig:FID}.
 
We are testing under the condition of $\textit{classifier-free} $ ${\textit{ guidance}}$ = 7.5, with DDIM 50 steps . This is a more common setting, although it does not give the best FID. It can be seen from Figure \ref{fig:FID} that the preloaded text-to-image capability has not been destroyed during the training process. And the FID value has been fluctuating around the FID value of the preloaded model during the training.

\subsection{Structure of Modules}
\begin{figure}
\begin{center}
    \includegraphics[width=0.45\textwidth]{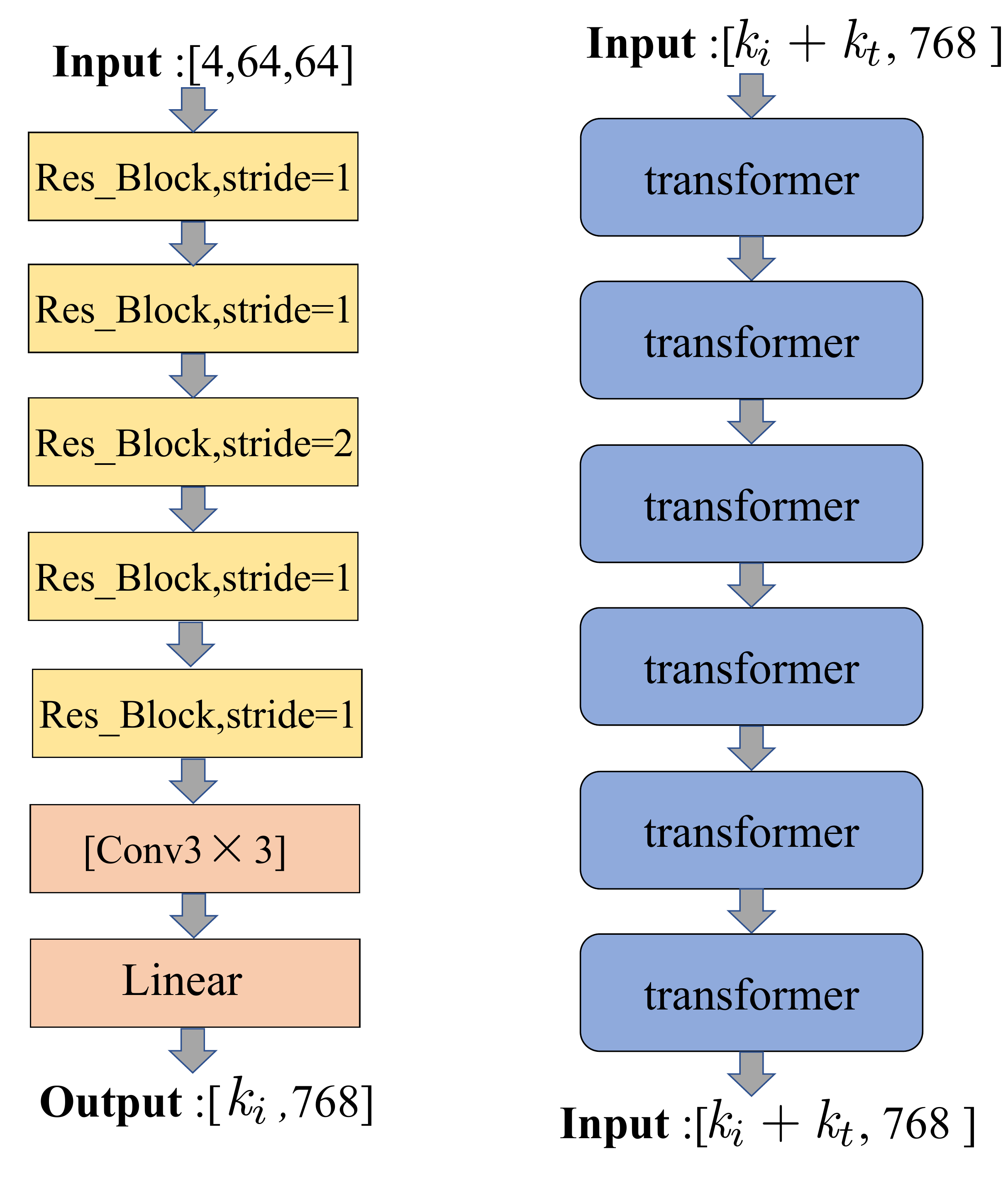}
    \caption{The structure of image encoder and transformer module.}
    \label{fig:structure}
\end{center}
\end{figure}

 An image encoder  and a transformer module are applied in our method. We design a simple image encoder for fast training and sampling, as shown in Figure \ref{fig:structure} (left). It consists of 5 \textit{Res\_Block}, a \textit{out\_layer} and a \textit{Linear}. In order to further reduce the consumption of computing resources,  the latent code, which is obtained by downsampling the input image with VAE, is used as the input for image encoder in  our experiments. The final image embedding dimension $\textit{k}_{i}$ can be controlled by the channels of \textit{out\_layer}. The transformer module, as shown in Figure \ref{fig:structure} (right), has 6 layers, 8 attention heads, 768 embedding dimensions and 3072 hidden dimensions. And the learnable positional embedding is also used. The length of the input sequence is the sum of the length of the text embedding ($\textit{k}_{t}$)and the length of the image embedding ($\textit{k}_{i}$). 
 
{\small
\bibliographystyle{ieee_fullname}
\bibliography{egbib}
}

\end{document}